\documentclass[a4paper,fleqn]{cas-dc}

\usepackage{cite}
\usepackage{amsmath,amssymb,amsfonts}
\usepackage{algorithmic}
\usepackage{textcomp}
\usepackage{float}
\usepackage{algorithm}
\usepackage{algorithmic}
\usepackage{comment}
\usepackage{makecell}
\usepackage{amssymb}
\usepackage{multirow}
\usepackage{caption}
\usepackage{subcaption}
\usepackage{xurl}
\usepackage{hyperref}
\hypersetup{breaklinks=true}

\usepackage{adjustbox}
\def\tsc#1{\csdef{#1}{\textsc{\lowercase{#1}}\xspace}}
\tsc{WGM}
\tsc{QE}

\begin{document}
\let\WriteBookmarks\relax
\def\floatpagepagefraction{1}
\def\textpagefraction{.001}

\shorttitle{}    

\shortauthors{}  

\title [mode = title]{DTC: A Deformable Transposed Convolution Module for Medical Image Segmentation}  


%

\author[1]{Chengkun Sun}



\ead{sun.chengkun@ufl.edu}




\author[1]{Jinqian Pan}


\ead{jinqianpan@ufl.edu}

\author[1]{Renjie Liang}


\ead{liang.renjie@ufl.edu}

\author[1]{Zhengkang Fan}


\ead{fanz@ufl.edu}

\author[2]{Xin Miao}


\ead{xin.miao.uta@gmail.com}

\author[3]{Jiang Bian}


\ead{bianji@iu.edu}





\author[1]{Jie Xu}

\cormark[1]
\ead{xujie@ufl.edu}
\cortext[cor1]{Corresponding author}
\affiliation[1]{organization={Department of Health Outcomes and Biomedical Informatics, University of Florida},
            city={Gainesville},
            postcode={32611}, 
            state={FL},
            country={USA}}
\affiliation[2]{organization={Tiktok},
            city={San Jose},
            postcode={95110}, 
            state={CA},
            country={USA}}
\affiliation[3]{organization={Regenstrief Institue, Indiana University, IU Health},
            city={Indianapolis},
            postcode={46202}, 
            state={IN},
            country={USA}}

\begin{abstract}
In medical image segmentation, particularly in UNet-like architectures, upsampling is primarily used to transform smaller feature maps into larger ones, enabling feature fusion between encoder and decoder features and supporting multi-scale prediction. Conventional upsampling methods, such as transposed convolution and linear interpolation, operate on fixed positions: transposed convolution applies kernel elements to predetermined pixel or voxel locations, while linear interpolation assigns values based on fixed coordinates in the original feature map. These fixed-position approaches may fail to capture structural information beyond predefined sampling positions and can lead to artifacts or loss of detail. Inspired by deformable convolutions, we propose a novel upsampling method, Deformable Transposed Convolution (DTC), which learns dynamic coordinates (i.e., sampling positions) to generate high-resolution feature maps for both 2D and 3D medical image segmentation tasks. Experiments on 3D (e.g., BTCV15) and 2D datasets (e.g., ISIC18, BUSI) demonstrate that DTC can be effectively integrated into existing medical image segmentation models, consistently improving the decoder’s feature reconstruction and detail recovery capability.
\end{abstract}


\maketitle

\section{Introduction}
\label{sec:introduction}

Medical image segmentation, a fundamental task in medical image analysis, involves dividing medical scans into anatomically or pathologically meaningful regions, thereby facilitating precise disease localization, quantitative assessment, and computer-assisted diagnosis~\cite{yu2023techniques,xu2024advances,abdou2022literature,galic2023machine,xu2025few}. In recent years, deep learning models have significantly advanced medical image segmentation, achieving state-of-the-art performance on major benchmarks and promoting continuous progress in clinical image understanding~\cite{minaee2021image,yu2023techniques}. The foundational deep learning model for segmentation, the fully convolutional network, introduced the architecture of feature extraction via downsampling followed by upsampling to restore spatial details~\cite{minaee2021image,long2015fully,huang2022fully}. Since then, models such as U-Net~\cite{ronneberger2015u}, based on an encoder–decoder architecture, subsequently,   UNETR~\cite{hatamizadeh2022unetr} was developed by the Vision Transformer~\cite{dosovitskiy2020image} and nnMamba~\cite{gong2024nnmamba} utilizing the Mamba architecture~\cite{gu2023mamba}, have all adopted this downsampling–upsampling paradigm for segmentation. These models have demonstrated strong performance across various modalities (e.g., CT, MRI, ultrasound) and dimensions (2D and 3D), contributing substantially to the adoption of deep learning in modern medical image segmentation~\cite{heimann2009comparison,clarke1995mri,noble2006ultrasound}. 

In addition to the aforementioned approaches that enhance the encoder by introducing Transformer or Mamba architectures, other studies have improved encoder performance through residual connections~\cite{he2016deep} or pool skip~\cite{sun2025beyond}. Most of these advances focus on improving the encoder, while fewer studies have explored methods to enhance the decoder in medical image segmentation models. 
In medical image segmentation models, the decoder plays a crucial role in reconstructing high-resolution predictions from compressed feature representations~\cite{azad2024medical}. Within the decoder, upsampling is a key operation for restoring spatial details and enabling multi-level feature fusion~\cite{wazir2025rethinking}. By aligning feature maps at different resolutions, upsampling enables feature fusion, blending low-level with high-level features~\cite{zhang2018exfuse}. This process enhances the model's ability of capturing spatial and contextual information across multiple layers, reducing local ambiguities~\cite{garcia2017review,badrinarayanan2017segnet}. Likewise, in multi-scale prediction, upsampling allows the network to generate multi-scale high-resolution feature maps directly from low-resolution ones, allowing it to handle objects of varying sizes and enhancing performance in tasks requiring precise localization~\cite{garcia2017review,roy2016multi}.

\begin{figure}
  \centering
   \includegraphics[width=0.45\textwidth]{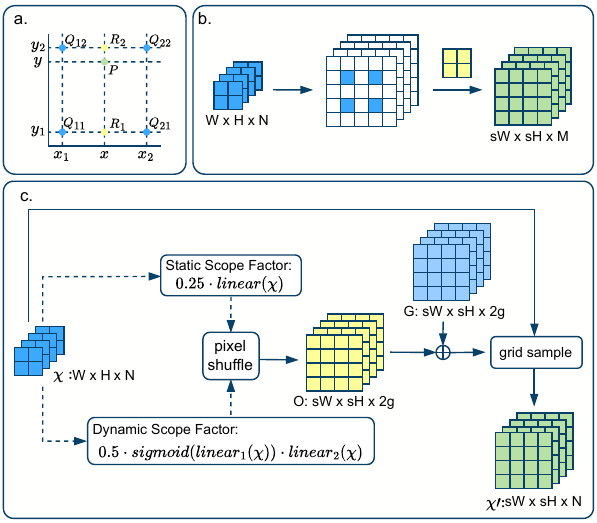}
   \caption{Three common upsampling structures: (a) linear interpolation, (b) transposed convolution (also known as deconvolution), and (c) Dysample. 
   Transposed convolution (b) allows channel adjustment by first zero-padding (indicated by white areas) and then convolving, while Dysample (c) maintains a consistent number of channels before and after upsampling. In Dysample, $g$ denotes the coordinate generation channel.}
   \label{fig:threeways}
\end{figure}

Two commonly used upsampling methods--linear interpolation and transposed convolution, as shown in Fig.~\ref{fig:threeways}a and b, transform small-sized feature maps into larger ones, regenerating information required for segmentation masks~\cite{minaee2021image}. Linear interpolation estimate values at upsampling points using a weighted combination of neighboring pixels or voxels from the original feature map~\cite{dumitrescu2019study}. In contrast, transposed convolution applies a learnable kernel to each small feature map, producing the pixel values for the upsampled map~\cite{dumitrescu2019study}. While both methods increase the spatial resolution of feature maps, they rely on fixed positions (fixed coordinates or pixel). These fixed-position approaches may not capture structural information beyond the predefined sampling positions~\cite{dolson2010upsampling}. 
Deformable convolution~\cite{dai2017deformable, zhu2019deformable, wang2023internimage, xiong2024efficient} addresses these limitations by allowing more flexible sampling of spatial features as shown in Fig.~\ref{fig:DCN}.
Dysample~\cite{liu2023learning}, a recent approach that incorporates learnable coordinates into interpolation, further improves upon linear interpolation, enhancing segmentation performance as shown in Fig.~\ref{fig:threeways}c.

Inspired by these advances, we introduced \textbf{Deformable Transposed Convolution (DTC)} (see Fig.~\ref{fig:DTC}), a novel upsampling structure that integrates learnable coordinates within the transposed convolution framework. This design enables dynamic and spatially adaptive upsampling, adjusting sampling positions according to learned offsets during the upsampling process. DTC can be integrated into different model architectures. We integrated DTC with CNN-based nnUNet~\cite{isensee2021nnu}, UNet~\cite{ronneberger2015u}, Transformer-based UNETR~\cite{hatamizadeh2022unetr}, SwinUNETR V2~\cite{he2023swinunetr}, and Mamba-based nnMamba~\cite{gong2024nnmamba}, SegMamba~\cite{xing2024segmamba}. Additionally, DTC was evaluated across multiple datasets, including the 2D medical image segmentation datasets (ISIC18~\cite{tschandl2018ham10000,codella2019skin}, BUSI~\cite{al2020dataset}), and the 3D medical image segmentation dataset (BTCV15~\cite{BTCV2015}). Experimental results indicate that DTC can improve segmentation performance across a variety of models and datasets.

\section{Related Work}
\label{sec:formatting}

\subsection{Deformable Convolution (DCN)}
\begin{figure}
  \centering
   \includegraphics[width=0.45\textwidth]{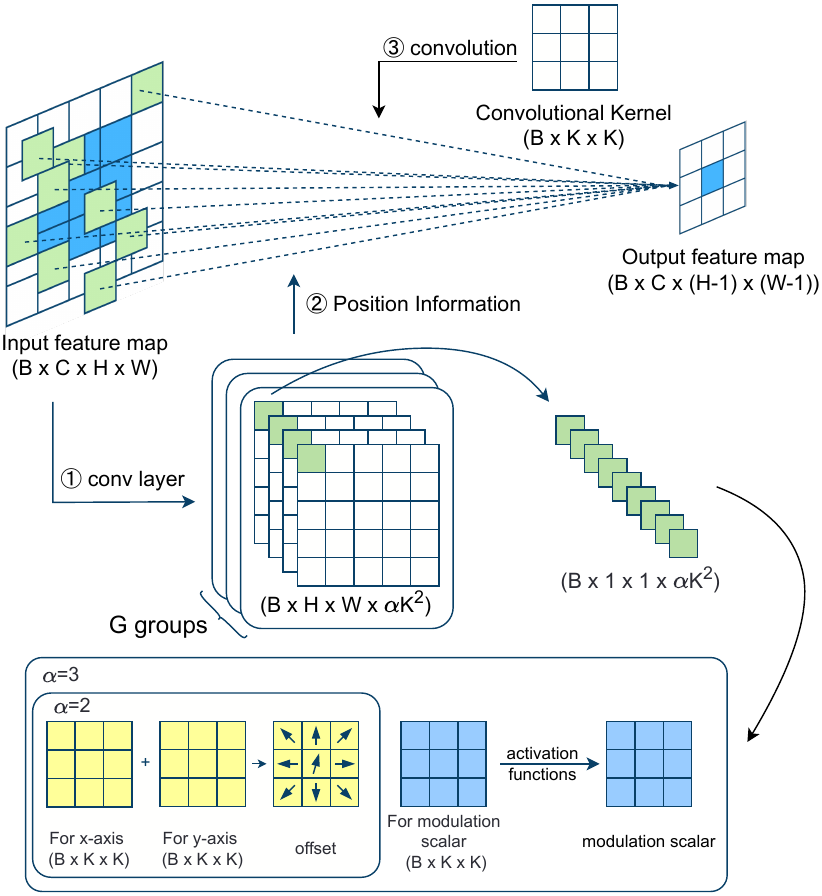}
   \caption{Structure of Deformable Convolution (DC). $\alpha$ represents the number of information groups required by the convolution kernel to generate coordinates. 
   K denotes the kernel size, B is the number of convolutions applied to the feature map, C is the number of feature channels, 
   and G is the number of input groups used in multi-head attention. 
   }
   \label{fig:DCN}
\end{figure}

Fig.~\ref{fig:DCN} illustrates the structure of various versions of deformable convolution.
Deformable convolution v1 (DCNv1) was first introduced in the deformable convolutional network~\cite{dai2017deformable} to address the limitations of traditional convolutional modules by learning dynamic offsets.
 These offsets are added to the input coordinates to generate new positions, enabling adaptive interpolation and creating a deformable feature map for further convolutional processing~\cite{dai2017deformable}.
To address the issue where features in DCNv1 may be influenced by irrelevant image content, DCNv2~\cite{zhu2019deformable} introduced modulation scalars.
These scalars, which are learned through convolution and range between 0 and 1, are applied to the product of the input element and the convolutional weight, allowing for better control over the influence of irrelevant areas. This enhances the flexibility and precision of the deformable convolution process, making it more effective for processing complex images. 

As a continuation of these advancements, DCNv3~\cite{wang2023internimage} introduced a multi-head mechanism to capture long-range dependencies. This extension allows the model to better account for spatial offsets, which can help in tasks requiring broader context. 
DCNv4~\cite{xiong2024efficient} improved efficiency by allowing a single head to process multiple channels simultaneously, reducing memory usage and simplifying bilinear interpolation.
Inspired by deformable convolution, we propose integrating deformable learning within the transposed convolution operation (see Fig.~\ref{fig:DTC}). This allows transposed convolutions to incorporate deformable capabilities, enabling adaptive sampling of relevant features during upsampling.

\subsection{Upsampling Methods}
Fig.~\ref{fig:threeways} shows three common upsampling structures.
Upsampling is a critical process in transforming low-resolution feature maps into high-resolution ones. A commonly used upsampling method is linear interpolation (see Fig.~\ref{fig:threeways}a), where new pixel or voxel values in the target feature maps are calculated as weighted averages of the neighboring pixels or voxels in the source image (1 $\times$ 1 for 2D, or 1 $\times$ 1 $\times$ 1 for 3D). 
While linear interpolation produces smooth upsampled images, its fixed coordinates placement limits its ability to adaptively learn spatial information. 

In contrast, transposed convolution is another popular upsampling technique that allows for learning (see Fig.~\ref{fig:threeways}b). Initially proposed for generating robust image representations~\cite{zeiler2010deconvolutional}, transposed convolution works by inserting zeros between each pixel in the feature map to expand the feature map and then applying a convolution to produce the new feature map. Therefore, the root of transposed convolution's approach to generating high-resolution images is the insertion of zero values. It still relies on fixed coordinates for upsampling, and the zero insertion process does not directly contribute to adaptive learning from the data.

To overcome these constraints, several dynamic upsampling techniques have been developed.  
CARAFE~\cite{wang2019carafe} directly uses convolution to learn a larger feature map by expanding the number of channels in the input features. Then, high resolution feature map pixels are supplemented through channel elements.
It generates new pixel values through a series of pointwise multiplication and concatenation operations. And the convolutional kernels are learned dynamically to adapt to input features. 
Building on this foundation, FADE~\cite{lu2022fade} and SAPA~\cite{lu2022sapa} further improve CARAFE's performance by fusing low-resolution and high-resolution information. 
However, the introduction of numerous convolutions and complex calculations significantly increases the model's size and the entire process does not involve coordinate transformation.

Another dynamic upsampling method is Dysample~\cite{liu2023learning}, which is designed based on point sampling, dividing a point into multiple points to achieve clearer edges (see Fig.~\ref{fig:threeways}c). 
It shares a similar concept with DC, as it generates a learnable coordinate through a scope factor, generated through a linear operation, followed by shuffle pixel processing, allowing for interpolation on the original image.  
Since this method relies on dynamic interpolation for upsampling, it has relatively few parameters. 
However, this method does not extract features from the original feature map via convolution, which may limit its performance compared to methods like FADE in some tasks~\cite{liu2023learning}. 
Additionally, transposed convolution allows adjusting the number of output feature channels, whereas FADE and Dysample maintain the same number of channels as the input.

\section{Deformable Transposed Convolution (DTC)}

We propose a novel deformable transposed convolution (DTC) to address the limitations of traditional transposed convolution, specifically its lack of learnable, dynamic upsampling capabilities due to the static insertion of zero values. Unlike other learnable upsampling methods, DTC retains the flexibility of traditional transposed convolution in adjusting channels and is compatible with other techniques like linear interpolation, FADE, and Dysample, adding a degree of learnability to the upsampling process.
\begin{figure*}
  \centering
   \includegraphics[width=0.8\textwidth]{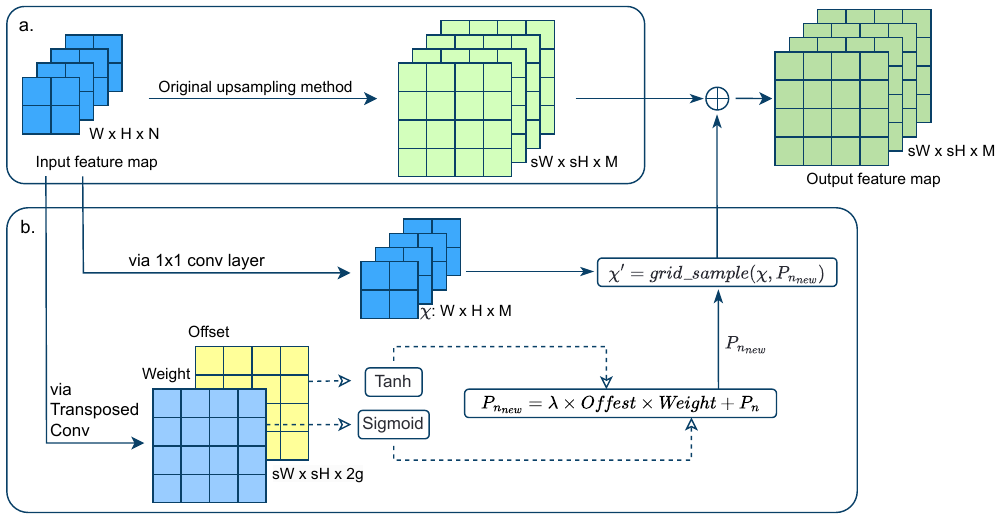}
   \caption{Structure of Deformable Transposed Convolution (DTC). The model consists of two components: (a) the upsampling method, typically implemented using linear interpolation or transposed convolution, and (b) the deformable transposed convolution component. In part (b), one path processes the feature information through convolution operations, while the other path generates offsets and weights via transposed convolution. These offsets and weight are used to compute new coordinates, $P_{n_{new}}$, which are combined with the convolution-generated features $\mathcal{X}$ through \textit{grid\_sample} to produce the new upsampled features, $\mathcal{X}'$. The outputs from both (a) and (b) are summed to form the final upsampling result. Here, W, H, N, and M represent the width, height, input channels, and output channels of the feature map, respectively. The upsampling scale is denoted by s, and $g$ represents the number of axes used for generating new coordinates through offsets and weights, with $g=2$ for 2D images and $g=3$ for 3D images. $P_{n}$ represents the original grid point coordinates, and $\lambda$ indicates the receptive field coefficient.}
   \label{fig:DTC}
   \vspace{-0.2cm}
\end{figure*}
\subsection{Deformable Strategies}
Fig.~\ref{fig:DTC} illustrates the entire process of the deformable convolution. Our approach reverses the traditional transposed convolution sequence by applying convolution before interpolation. This preserves transposed convolution's strengths in feature extraction and channel transformation while enabling a deformable interpolation effect through dynamic, learnable coordinates. 
DTC consists of two main components: the convolution processing component and the coordinate generation component.

In the convolution processing component, a $1 \times 1$ convolution is applied to input features, retaining transposed convolution's feature extraction capability while facilitating channel adjustment. The coordinate generation component produces dynamic offsets and weights via a transposed convolution layer, using four channels for 2D images (two for x and y offsets, two for weight control) and six channels for 3D images. 
The tanh and sigmoid activations constrain offsets to the range [-1, 1] and weights to [0, 1], respectively. 
The new learnable coordinates, $P_{n_{new}}$, are generated using Eq.~(\ref{coordinate}), where $P_{n_{Grid}}$ represents grid coordinates for linear interpolation, and $\lambda$ is a receptive field hyperparameter. 
\begin{equation}
P_{n_{new}} = \lambda \times \textit{Offset} \times \textit{Weight} + P_{n_{Grid}}.
\label{coordinate}
\end{equation}
Finally, we apply grid\_sample to the obtained features $\mathcal{X}$ and the new coordinates $P_{n_{new}}$ to achieve the upsampling result.
\begin{equation}
\mathcal{X} ' = grid\_sample(\mathcal{X},P_{n_{new}}).
\label{grid_sample}
\end{equation}

\subsection{Receptive Field}
In other deformable approaches, coordinates can shift across the entire feature map, giving these methods a fully unrestricted receptive field. This design potentially addresses the limitation of traditional upsampling methods, which often lack long-range dependency. However, a fully unrestricted receptive field may not always be optimal, as excessive freedom could reduce stability or precision during learning. Thus, we introduce a controlled receptive field in our proposed DTC. 

To achieve this, we add a hyperparameter, $\lambda$, which scales the offset to control the maximum distance a single coordinate can shift. The value of $\lambda$ is calculated as $1$ divided by the size of the input feature map, providing the ratio of pixel width to the coordinate range (which spans from -1 to 1 for the downsampled coordinates). Given that the offset-weight product is also constrained between -1 and 1, scaling by $\lambda$ allows precise control over the range of sampling coordinates. Since the initial input features undergo a $1 \times 1$ convolution, the base receptive field is effectively 2, making the receptive field of the entire deformable transposed convolution fully dependent on the coordinate movement range. Adjusting the value of $\lambda$ enables flexible control over the receptive field size, with the default set to one pixel-width of the feature map, giving the deformable transposed convolution a base receptive field of 2. The impact of this parameter will be explored in our ablation experiments. Finally, we use the generated coordinates to perform linear interpolation on the feature map obtained from the convolution, producing the deformable transposed convolution's upsampling output.

\subsection{Comparison with Other Deformable Mechanisms}
Our DTC approach diverges from other deformable methods like Dysample by avoiding pixel shuffling and complex multiplication-addition processes. Instead, it builds on DCNv1 and DCNv2 by generating a single transposed convolution operation to yield coordinate offsets and weights. While DCNv2 uses modulation scalars to adjust coordinates, our method applies a weight to each coordinate, providing additional control. 
Ablation experiments demonstrate the effectiveness of this structure. In our method, coordinates are generated in a single step using transposed convolution, while deformable convolution performs deformation for each convolution operation. 

Additionally, in DCNv3 and DCNv4, the coordinate values required for each sampled pixel depend on the size of the convolution kernel. For example, a $3\times3$ kernel requires 9 coordinate points. Due to the need for multiple convolution operations on features, multi-dimensional tensors are included, each containing 9 heads, making it suitable to use a multi-head mechanism. Then a multi-head mechanism is typically applied with a default setting of 9 heads. In contrast, our method performs upsampling only once instead of multiple times, so multi-head attention cannot be directly applied.

DTC's design complements existing upsampling methods by providing adaptive upsampling while maintaining feature details. By fusing the DTC module’s deformable upsampling with the model’s original upsampling results (e.g., via linear interpolation or traditional transposed convolution), we achieve an upsampling outcome that balances prominent feature focus with fine detail preservation. 
\begin{table*}[t]
    \scriptsize
    \centering
    \caption{Comparison of segmentation results on the ISIC and BUSI datasets by incorporating different upsampling methods into the UNet model. 
    UNet is evaluated with two common upsampling methods: bilinear interpolation (indicated as $_{linear}$) and transposed convolutions (indicated as $_{tc}$). }
    \begin{subtable}[t]{0.48\textwidth}
        \centering
        
        \small
        \begin{tabular}{l|r||l|r}
        \hline
            \textbf{Models}    & DICE (\%)    & \textbf{Models}    & DICE (\%)              \\  \hline
            UNet$_{linear}$     &   78.23$\pm$2.38         &  UNet$_{tc}$     & 78.57$\pm$1.08       \\  
              +Dysample   &  79.29$\pm$1.55         &   +Dysample   &  --       \\ 
             +FADE  &  78.80$\pm$1.45              &   +FADE       &   -- \\ 
             +DTC   & \textbf{79.58$\pm$1.47}   &    +DTC   & \textbf{79.41$\pm$1.02}   \\ 
            \hline
        \end{tabular}
        \caption{Results on the ISIC dataset.}
        \label{tab:busi_a}
    \end{subtable}
    \hfill
    \begin{subtable}[t]{0.48\textwidth}
        \centering
        \small
        \begin{tabular}{l|r||l|r}
        \hline
            \textbf{Models}    & DICE (\%)    & \textbf{Models}    & DICE (\%)              \\  \hline
            UNet$_{linear}$     &   75.08$\pm$1.76         &  UNet$_{tc}$     & 75.76$\pm$1.24        \\  
              +Dysample   &  76.24$\pm$1.14         &   +Dysample   &  --       \\ 
              +FADE  &  75.80$\pm$2.02              &  +FADE       &   -- \\ 
              +DTC   & \textbf{78.14$\pm$1.69}   &    +DTC   & \textbf{78.49$\pm$1.63}   \\ 
            \hline
        \end{tabular}
        \caption{Results on the BUSI dataset.}
        \label{tab:busi_b}
    \end{subtable}

    \label{tab:main_busi}
\end{table*}
\section{Experiments}
\subsection{2D Medical Image Segmentation}
\paragraph{Datasets}We used the International Skin Imaging Collaboration (ISIC) dataset~\cite{tschandl2018ham10000,codella2019skin} for skin lesion segmentation. The ISIC 2018 challenge, held at the MICCAI conference, comprised three tasks and included over 12,500 dermoscopic images. Additionally, we employed the Breast Ultrasound Images (BUSI) dataset~\cite{al2020dataset}, a publicly available collection designed for research on breast lesion detection and classification. BUSI contains 780 images collected from 600 female patients, with annotations provided by experienced radiologists. The dataset covers three categories: normal tissue, benign lesions, and malignant lesions, with corresponding segmentation masks for the lesion regions.

\paragraph{Compared Methods} 
We evaluated the DTC across U-Net~\cite{ronneberger2015u}, while ``+DTC'' models were trained with our proposed block. The reason for selecting this model is that the UNet architecture is the most widely used baseline in medical image segmentation. Its encoder incorporates multiple upsampling units, and its two configurations—one utilizing linear interpolation and the other using transposed convolution—allow for an effective comparison of the proposed DTC performance against that of other modules. In addition, we selected two recent and efficient 2D segmentation models, including the Transformer-based SwinUNETR V2~\cite{he2023swinunetr} and the Mamba-based SegMamba~\cite{xing2024segmamba}, to verify that the proposed DTC module can adapt to different encoder architectures. We conducted 2D experiments to compare different advanced upsampling methods, namely FADE~\cite{lu2022fade} and Dysample~\cite{liu2023learning}, by replacing the original upsampling layers in UNet and examining their performance relative to DTC. 
Notably, Dysample represents a recent state-of-the-art upsampling technique, while FADE, proposed in the Dysample paper, has been shown to outperform Dysample on certain datasets. 
These comparisons allow a more comprehensive evaluation of DTC's effectiveness against other leading upsampling approaches.

\begin{table}[htbp]
    \centering
    \caption{Comparison of segmentation performance on the ISIC and BUSI datasets using SegMamba and SwinUNETR V2, with and without the proposed DTC module.}
    \small
    \begin{tabular}{l|r|r}
    \hline
          \textbf{Models}   & \makecell{ISIC \\ DICE (\%)} & \makecell{BUSI \\ DICE (\%)} \\ \hline 
        SegMamba     &  78.21$\pm$0.91        & 75.43$\pm$1.78      \\  
        \quad + DTC   & \textbf{78.70$\pm$1.42}          &  \textbf{76.66$\pm$1.51}      \\ 
        SwinUNETR V2  &  77.18$\pm$1.23             & 76.02$\pm$1.18 \\ 
        \quad + DTC   & \textbf{77.40$\pm$1.14}    & \textbf{76.35$\pm$2.22}   \\ 
        \hline
    \end{tabular}
    \label{results_2d_segmentaion_model}%
\end{table}

\paragraph{Experimental Settings}
We utilized the training, validation, and test datasets provided by the ISIC 2018 challenge. These datasets were combined and then randomly split into training and testing sets in a 5:2 ratio (2,600 images for training and 1,094 for testing). In addition, we used the publicly available BUSI breast ultrasound dataset, which includes images of benign and malignant breast tumors. A total of 87 images were randomly selected as the test set, and 520 images were used for training. We performed 5-fold cross-validation, selecting the optimal model from each fold's validation set. The selected models were then evaluated on the testing set, and we recorded the mean and variance of performance metrics across the 5 folds.

For data augmentation, we normalized pixel values and resized the images to $256 \times 256$ to meet the input requirements of the proposed block. This basic augmentation aimed to enhance the robustness of the models during training. The models were trained using the AdamW~\cite{loshchilov2017decoupled} optimizer with a weight decay of $1e-5$ and a learning rate of $1e-4$. Each model underwent 10,000 iterations of training, with the goal of achieving the highest DICE scores~\cite{milletari2016v}. 

\begin{figure*}
  \centering
\includegraphics[width=\textwidth]{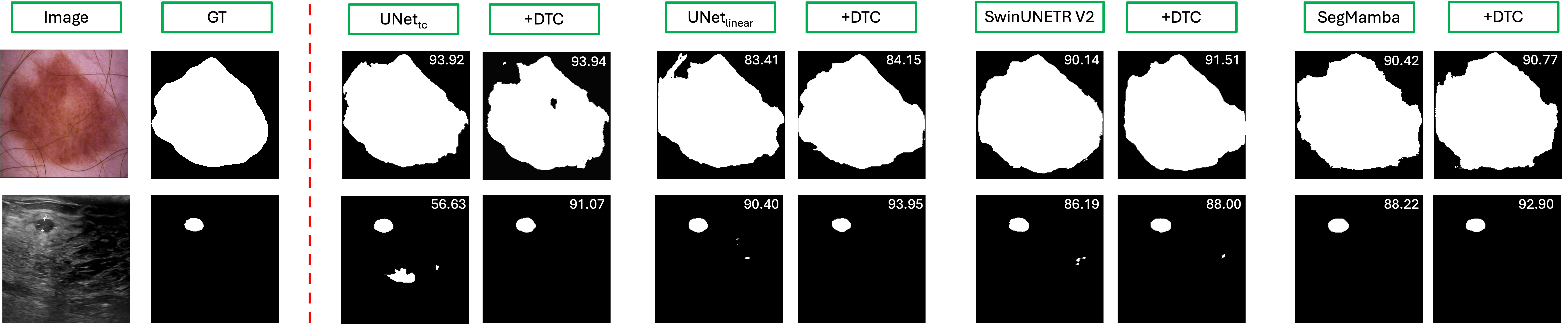}
   \caption{Example segmentation results on the ISIC (top row) and BUSI (bottom row) datasets. Each column corresponds to a different method, with white regions indicating the predicted segmentation masks. The numbers denote the Dice scores (\%) for each prediction. Results also show the segmentation performance when DTC is introduced into different models.}
   \label{vis_2d_deform}
\end{figure*}

\paragraph{Experimental Results}
Table \ref{tab:main_busi} presents the DICE improvements obtained by adding DTC to UNet. On the ISIC dataset, DICE increased by +1.35\% for bilinear UNet and +0.84\% for transposed UNet. On the BUSI dataset, DICE improved by +3.06\% for bilinear UNet and +2.73\% for transposed UNet, suggesting that DTC can improve segmentation performance across multiple datasets and upsampling strategies. Table \ref{results_2d_segmentaion_model} compares the segmentation performance of SegMamba and SwinUNETR V2 on the ISIC and BUSI datasets, with and without the proposed DTC module. For SegMamba, DICE improved by +0.49\% on ISIC and +1.23\% on BUSI. For SwinUNETR V2, DICE increased by +0.22\% on ISIC and +0.33\% on BUSI, demonstrating that integrating DTC consistently enhances segmentation accuracy across different models and datasets. It should be noted that, for the experiments using DTC, the receptive field was adjusted differently for each model. 
For SegMamba, the receptive field was set to 4 for ISIC and 10 for BUSI; for SwinUNETR V2, it was set to 2 for ISIC and 4 for BUSI. For UNet$_{tc}$, the receptive field was set to 10 for both datasets, 
while for UNet$_{linear}$, it was set to 10 for ISIC and considered infinite for BUSI. 
The impact of different receptive field settings will be further analyzed in the subsequent analysis.
\begin{figure}
  \centering
   \includegraphics[width=0.3\textwidth]{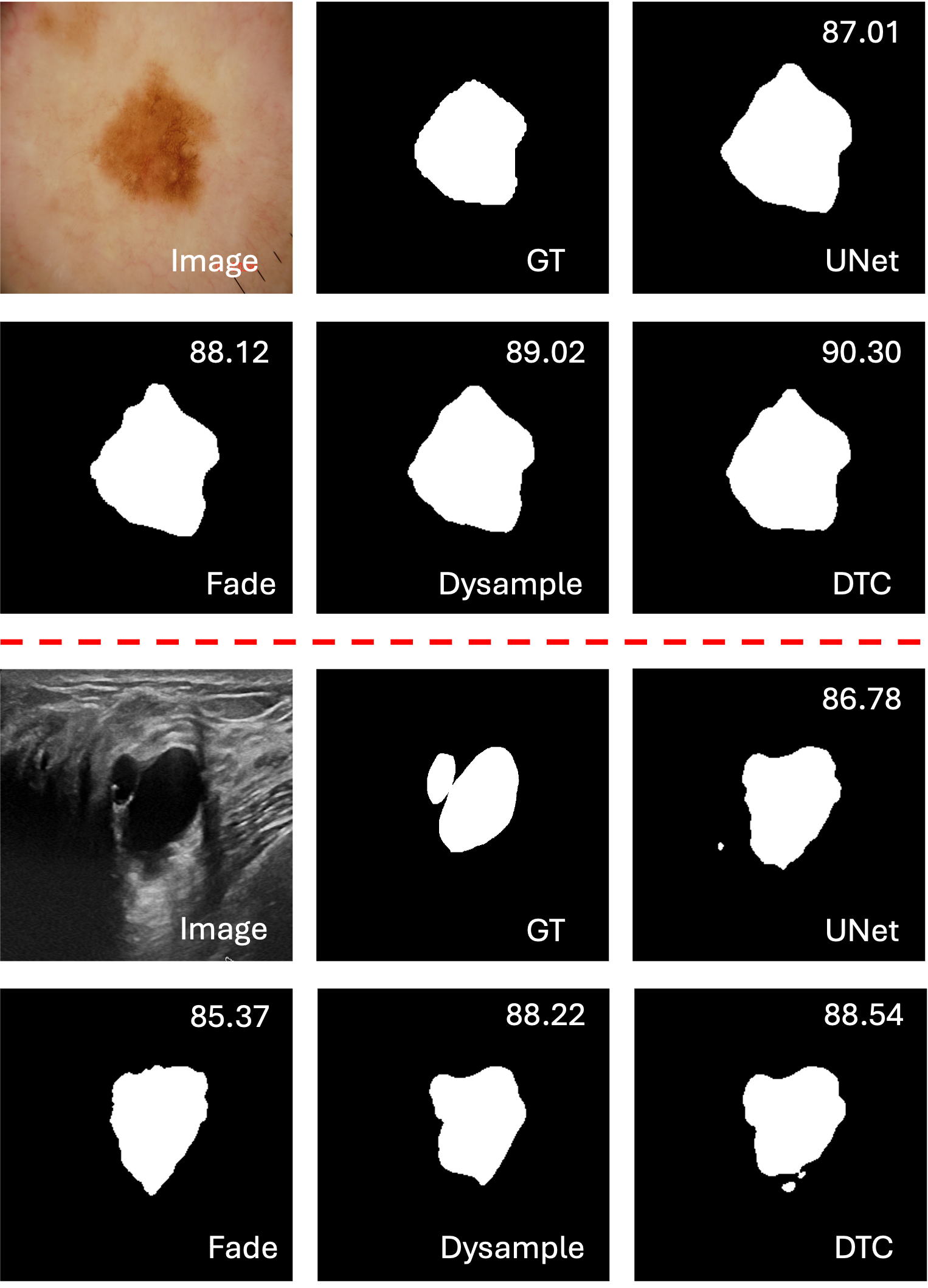}
   \caption{Visualization comparing the effects of different upsampling methods on the same UNet model. The labels below each image indicate the model or the type of upsampling used, and the Dice score (\%) for each segmentation is shown at the top-right corner.}
   \label{vis_2d_upsample}
\end{figure}
Table~\ref{tab:main_busi} summarizes the performance gains of different upsampling methods applied to UNet$_{linear}$ on the ISIC and BUSI datasets. 
On ISIC, DICE increased by $+1.35\%$ with DTC, showing that DTC outperforms the second-best method (Dysample) by $+0.29\%$. 
On BUSI, the corresponding improvements were  $+3.06\%$, with DTC surpassing Dysample by $+1.90\%$. These results demonstrate that while all advanced upsampling techniques enhance segmentation performance, DTC consistently achieves the largest gains across datasets. For UNet$_{tc}$, the introduction of DTC improved the DICE score by $+0.84\%$ on ISIC and $+2.73\%$ on BUSI. Fig.~\ref{vis_2d_upsample}, particularly for the BUSI segmentation, indicates that DTC is more effective than other transposed convolution methods in suppressing background noise. From the Fig.\ref{vis_2d_deform}, it can be observed that the introduction of DTC improves the model’s ability to suppress background noise. On the other hand, after introducing DTC, the model becomes more sensitive to edge gradient information. For example, in a UNet with transposed convolution, DTC helps eliminate background artifacts such as hair in ISIC segmentation. However, the normally segmented light-colored region in the upper-left corner is not captured; instead, only the dark-colored region is segmented.

\begin{figure*}[htbp]
  \centering
\includegraphics[width=\textwidth]{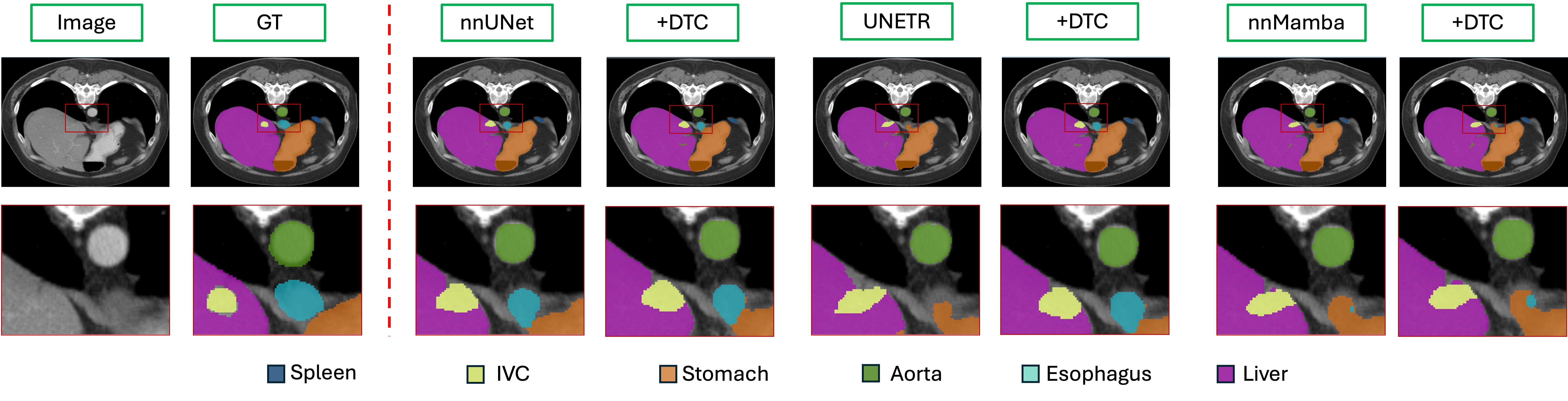}
   \caption{Visualization of the DTC module applied to a 3D segmentation model. The first row shows the 2D segmentation slices, and the second row presents the zoomed-in views of the regions highlighted by the red rectangles. “IVC” is the abbreviation for Inferior Vena Cava.}
   \label{vis_discussion}
\end{figure*}

\subsection{3D Medical Image Segmentation}
\paragraph{Dataset} 
For the BTCV dataset, 12 scans were assigned to the test set, and 18 to the training and validation set. 
We performed 5-fold cross-validation on all models, averaging their softmax outputs across folds to determine voxel probabilities. Evaluation metrics included DICE~\cite{milletari2016v} and Normalized Surface DICE (NSD)~\cite{nikolov2018deep}.
\paragraph{Comparison Methods} 
We evaluated the DTC using the CNN-based model nnUNet~\cite{isensee2021nnu}, the Transformer-based model UNETR~\cite{hatamizadeh2022unetr} and the Mamba-based nnMamba~\cite{gong2024nnmamba}. In structure setup, nnUNet and UNETR use transposed convolution for upsampling, while nnMamba uses trilinear interpolation for the same purpose. For nnUNet, our implementation closely follows the nnUNet framework~\cite{isensee2021nnu}, covering data preprocessing, augmentation, model training, and inference. Scans and labels were resampled to the same spacing as recommended by nnUNet. We excluded nnUNet's post-processing steps to focus on evaluating the model's core segmentation performance. For a fair comparison, when reproducing nnUNet, we retained its default configuration. For the other models, we adopted the preprocessing settings consistent with nnUNet. On the other hand, our UNETR and nnMamba models were not trained using deep supervision.
\begin{table}[H]
    \centering
    \caption{Comparison of segmentation results on the BTCV dataset.}
    \small
    \begin{tabular}{l|l|l}
    \hline 
          \textbf{Models}   & DICE (\%) & NSD (\%) \\ \hline 
        nnUNet   &  81.52 & 76.10        \\  
        \quad + DTC   & \textbf{81.66 \  \ +0.14} & \textbf{75.97 \  \ -0.13}    \\ \hline
        UNETR  &   69.65 & 59.18  \\ 
        \quad + DTC   & \textbf{71.98 \  \ +2.33}  & \textbf{62.15 \  \ +2.97}  \\ \hline
        nnMamba   & 75.47 & 67.05   \\ 
        \quad + DTC   & \textbf{76.77 \  \ +1.30}  & \textbf{68.19 \  \ +1.14}  \\ 
        \hline
    \end{tabular}

    \label{tab:results_3d}%
\end{table}
\paragraph{Experimental Results}

The experimental results in Table~\ref{tab:results_3d} show that DTC provides notable performance improvements for UNETR and nnMamba, while the performance improvement for nnUNet is relatively smaller. This difference may be due to nnUNet not applying a uniform upsampling stride across all dimensions, resulting in limited learning of 3D features in some dimensions. Consequently, the 3D upsampling process approximates a combination of 2D results, which may constrain the deformable structure’s ability to capture full 3D information. This constraint prevents our coordinate generation process from fully leveraging DTC's potential for enhancing performance. In the Fig.~\ref{vis_discussion}, the esophagus was not segmented by UNETR and nnMamba; however, after introducing DTC, it is successfully segmented.

\subsection{Further Analysis}
\paragraph{Methods of Coordinate Generation}

We conducted ablation experiments on the coordinate generation module across four configurations, using UNet as the base model and applying bilinear interpolation and transposed convolution for upsampling within the ISIC experimental setup. The primary goal of these experiments was to assess the necessity of incorporating weights and using an activation function to constrain the range of coordinates and weights, as summarized in Table~\ref{tab:coordanate}. Results indicate that the coordinate generation approach outlined in Eq.~(\ref{coordinate}) can improve performance in both bilinear interpolation and transposed convolution methods.
\begin{table}
    \caption{The ablation experiments of coordinate generation on the BUSI dataset. The checkmark indicates that the model has adopted that particular component.}
    \centering
    \small
      \begin{tabular}{p{0.6cm}p{0.8cm}p{0.6cm}p{0.5cm}|p{1.5cm}|p{1.5cm}}
      \hline
       &  &  &   & \multicolumn{2}{c}{DICE (\%)}  \\ \cline{5-6}
      Weight & Sigmoid & Offset & Tanh  & UNet$_{linear}$  &  UNet$_{tc}$ \\
      \cline{1-4}\cline{5-6}
            &       & \checkmark     &       & 62.32$\pm$3.68 & 70.54$\pm$2.55 \\
            &       & \checkmark     & \checkmark     & 69.82$\pm$2.19 & 78.06$\pm$1.08 \\
      \checkmark     &       & \checkmark     &       & 56.31$\pm$4.92 & 49.70$\pm$5.78 \\
      \checkmark     &       & \checkmark     & \checkmark     & 56.59$\pm$4.57 & 55.91$\pm$4.11 \\
      \checkmark     & \checkmark     & \checkmark     &       & 72.59$\pm$3.55 & 74.50$\pm$2.96 \\
      \checkmark    & \checkmark     & \checkmark     & \checkmark     & \textbf{78.14$\pm$1.69} & \textbf{78.49$\pm$0.69} \\
      \hline
      \end{tabular}%
    \label{tab:coordanate}%
  \end{table}%
\begin{figure*}[htbp]
  \centering
\includegraphics[width=0.8\textwidth]{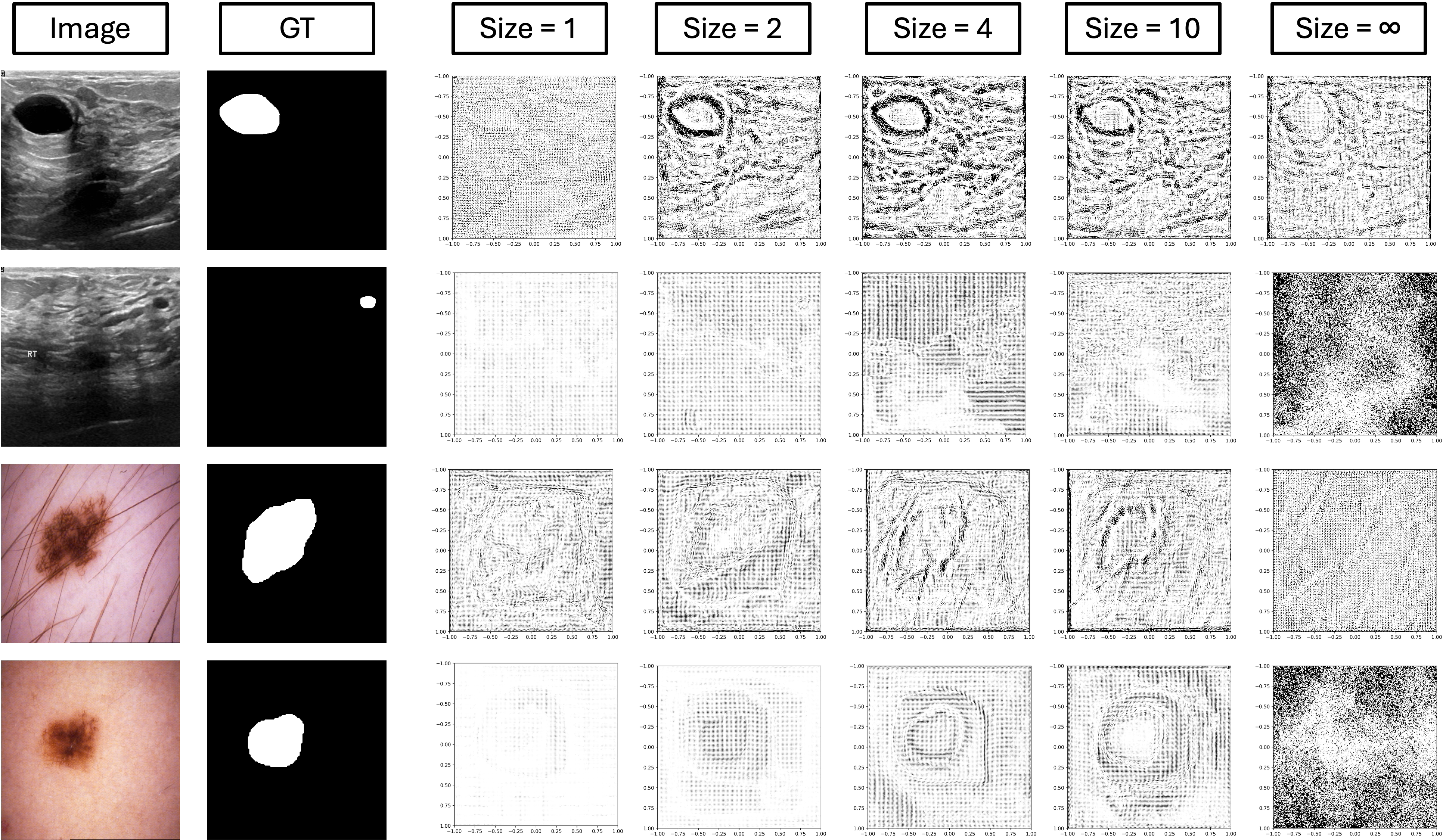}
   \caption{Visualization of coordinate scatter distribution in the last upsampling layer of UNet. 
   Top row: UNet with DTC using transposed convolution; bottom row: DTC with bilinear interpolation.
   White points indicate focused regions during upsampling, while black areas show filtered-out regions, highlighting the model’s targeted feature extraction.
   }
   \label{visualization}
\end{figure*}
\begin{figure*}[htbp]
  \centering
\includegraphics[width=0.85\textwidth]{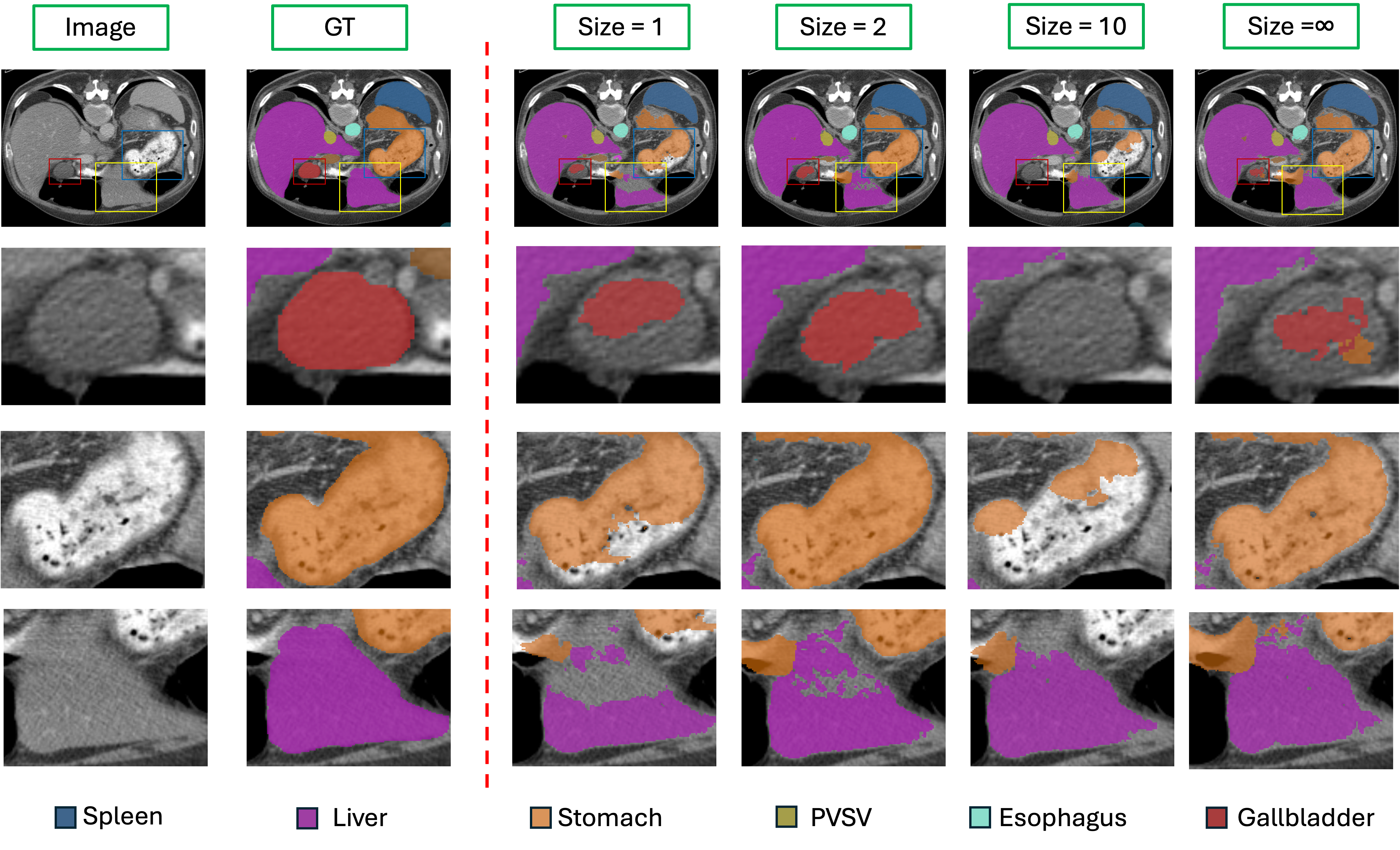}
   \caption{Effects of DTC with different receptive fields on the UNETR model. The first row shows a 2D slice of the 3D CT segmentation results. The second, third, and fourth rows display zoomed-in views of the regions highlighted by red, blue, and yellow rectangles, respectively. PVSV stands for Portal Vein and Splenic Vein.}
   \label{vis_3d_cor}
\end{figure*}

\paragraph{Receptive Field}
We tested the impact of receptive field size on model performance using the 2D ISIC dataset and the 3D BTCV dataset, evaluating both linear interpolation and transposed convolution models. Results, shown in Table~\ref{tab:receptive}, indicate that the DTC method is particularly sensitive to receptive field size. When the receptive field exceeds 10, model performance declines notably. While UNETR achieves high DICE scores with a full receptive field, it exhibits low NSD performance, suggesting that an overly large receptive field may reduce the model’s precision in capturing segmentation boundaries. Therefore, a smaller receptive field may improve boundary detection, but the default receptive field of 2 is not always optimal. Fine-tuning this receptive field yields better results. On the other hand, we performed a visualization analysis (Fig.~\ref{vis_3d_cor}) of the results for the UNETR model. 
We observed that adjusting the receptive field can improve overall model performance. 
However, in multi-class segmentation tasks, DTC does not consistently yield significant improvements for every organ and may result in lower segmentation performance for individual organs. 
For example, as shown in the (Fig.~\ref{vis_3d_cor}, segmentation of the stomach and gallbladder is clearly improved at the optimal receptive field, whereas the liver's performance deteriorates compared to using a receptive field of 10 or the full image. 
These observations suggest that the effectiveness of DTC may vary across organs in multi-class scenarios, highlighting the importance of organ-specific considerations when applying advanced upsampling techniques.

\begin{table}[htbp]
    \caption{The impact of receptive field size on segmentation performance. The 3D segmentation models nnMamba and UNETR were evaluated using the BTCV dataset. The 2D segmentation models, UNet$_{TC}$ and UNet$_{linear}$, were conducted on the ISIC dataset.}
  \centering
  \small
    \begin{tabular}{c|c|c|c|c|c}
    \hline
    \multicolumn{1}{c|}{\multirow{2}[4]{*}{\textbf{Models}}} &       & \multicolumn{4}{c}{Receptive Field} \\
\cline{3-6}          &       & $\infty$     & 10     & 2     & 1 \\
    \hline
    \multicolumn{1}{c|}{\multirow{2}[2]{*}{nnMamba}} & DICE (\%)  & 75.48     & 75.46      & 75.56      & \textbf{76.77} \\
          & NSD (\%)   & 66.86     & 66.84     & 67.12    & \textbf{68.19} \\
    \hline
    \multicolumn{1}{c|}{\multirow{2}[2]{*}{UNETR}} & DICE (\%)  & 71.94     & 70.99     & \textbf{71.98}     & 69.39 \\
          & NSD (\%)   & 61.76     & 61.78     & \textbf{62.15}     & 59.13 \\
    \hline
    UNet$_{TC}$  & DICE (\%)   & 70.85     & \textbf{79.41}    & 78.44     & 77.65 \\
    \hline
    UNet$_{linear}$  & DICE (\%)   & 77.30     & \textbf{79.58}     & 78.83     & 78.93 \\
    \hline
    \end{tabular}%

  \label{tab:receptive}%
\end{table}%

To further analyze, we visualized the coordinates in the last layer of the transposed coder during the segmentation process of the 2D UNet model, (see Fig.~\ref{visualization}). The proposed DTC is integrated into the model's original convolutional layers, serving as a complementary module to the existing convolutions. The visualization results indicate that the deformable module can complement conventional upsampling methods, such as transposed convolution or interpolation, by capturing information beyond the immediate edges of the segmentation target. 
This allows standard upsampling techniques to better focus on boundary regions. 
However, for small targets, as illustrated in the second column of the examples, DTC does not provide a strong improvement in feature attention. Furthermore, changes in the receptive field may reduce the ability of standard upsampling methods to emphasize boundaries, highlighting the sensitivity of edge-focused performance to receptive field settings.

\begin{table}[htbp]
    \caption{DICE (\%) results showing compatibility with different upsampling methods. The baseline is UNet.}
  \centering
  \small
    \begin{tabular}{l|ccc}
    \hline
    \textbf{Models} & ISIC & BUSI\\
    \hline
    Dysample & 76.24$\pm$1.14 & 79.29$\pm$1.55 \\
    \hline
     \ \ \ + DTC & \textbf{78.72$\pm$0.71} & \textbf{80.49$\pm$0.30}  \\
    \hline
    FADE  & 75.80$\pm$2.02 & \textbf{78.80$\pm$1.45} \\
    \hline
     \ \ \ + DTC & \textbf{76.87$\pm$1.41} & 78.64$\pm$1.05 \\
    \hline
    \end{tabular}%

  \label{tab:campatibility}%
\end{table}%
\paragraph{Compatibility with Other Upsampling}

\begin{figure}
  \centering
   \includegraphics[width=0.45\textwidth]{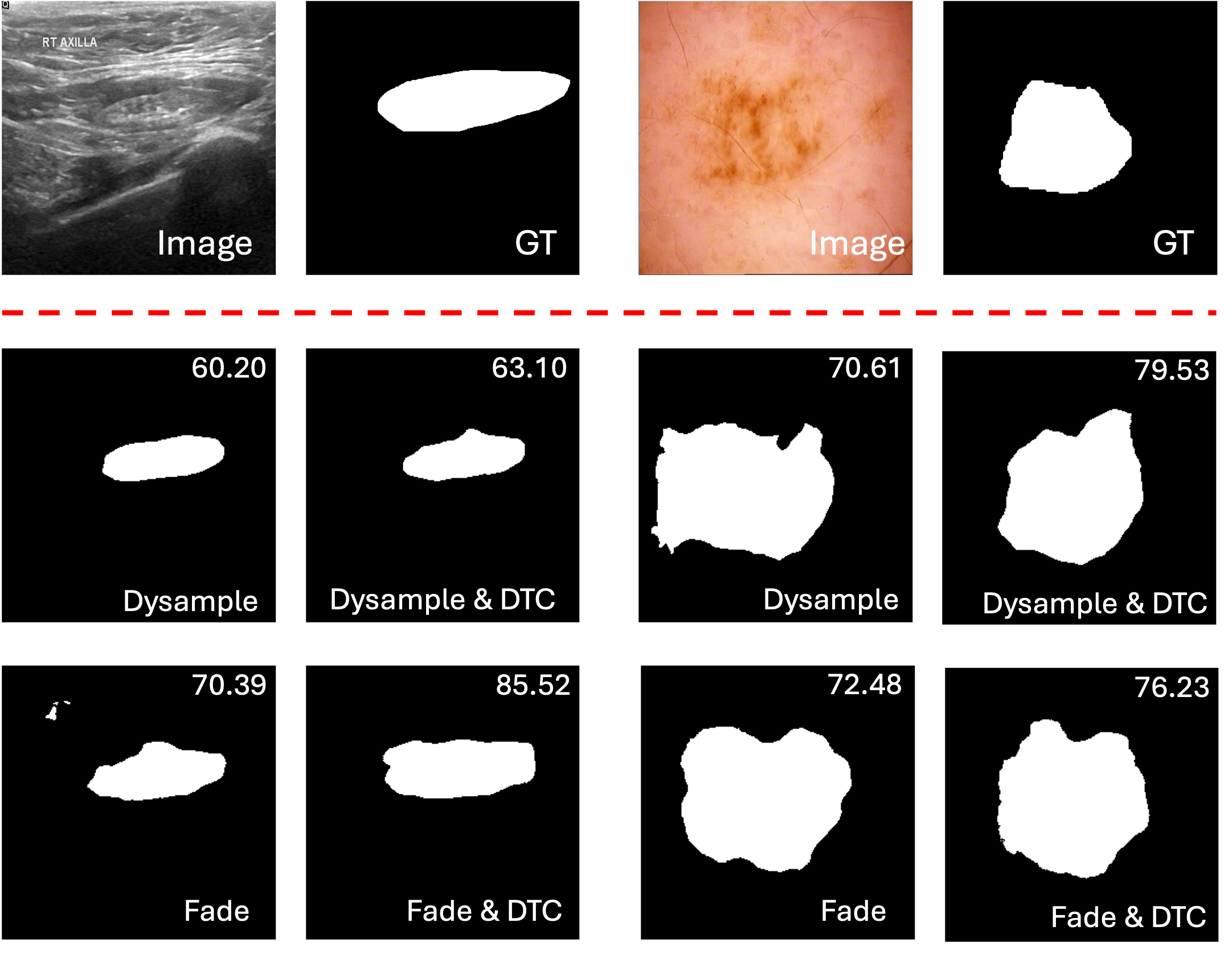}
   \caption{ Segmentation results with different upsampling methods, illustrating the compatibility of DTC with other types of upsampling. The first row shows the input images and corresponding ground truth labels. The second and third rows display the segmentation outputs from the models, with the type of upsampling method used in the UNet indicated below each image and the Dice score (\%) for each prediction shown at the top-right corner.}
   \label{vis_2d_campatible}
\end{figure}

Table~\ref{tab:campatibility} summarizes the Dice (\%) improvements demonstrating the compatibility of DTC with different upsampling methods on the ISIC and BUSI datasets. Incorporating DTC into Dysample results in a Dice gain of +2.48 on ISIC and +1.20 on BUSI. For FADE, adding DTC increases the Dice by +1.07 on ISIC, while showing a slight decrease of -0.16 on BUSI, indicating that DTC can be effectively integrated with other convolutional operations and that combining DTC with Dysample may lead to better segmentation performance. Fig.~\ref{vis_2d_campatible} illustrates the improvements brought by DTC for different upsampling methods. It can be observed that introducing DTC effectively suppresses background noise.


\paragraph{Computational Load}
Compared to the original UNet model, our approach introduces only a modest increase in parameters, adding just 1.3 million to the original 65.95 million, as shown in  Table~\ref{tab:parameter}.
Despite this slight increase, our method demonstrates better compatibility and stability over other state-of-the-art upsampling techniques.
\begin{table}[htbp]
     \caption{Computational load of various upsampling methods based on UNet$_{linear}$ model.}
  \centering
  \small
    \begin{tabular}{l|c|c|c|c}
    \hline
    \textbf{Models} & \multicolumn{1}{l|}{UNet} & \multicolumn{1}{l|}{+ Dysample} & \multicolumn{1}{l|}{+ Fade} & \multicolumn{1}{l}{+ DTC} \\
    \hline
    Params (M) & 65.95 & 65.97 & 66.55 & 67.25 \\
    \hline
    FLOPs (G) & 40.13 & 40.18 & 42.32 & 40.52 \\
    \hline
    \end{tabular}%
  \label{tab:parameter}%
\end{table}%

\section{Discussion}
Our experiments demonstrate that deformable transposed convolution (DTC) can effectively enhance segmentation model performance across both 2D and 3D tasks, including segmentation with and without background images. However, this approach has certain limitations that suggest directions for future work.

First, while DTC improves segmentation by reducing noise, this process may unintentionally exclude some image details during sampling. When local receptive fields contain uniformly important but noisy information, the generated coordinates may force the model to focus on background noise rather than suppressing it. To mitigate this, retaining the original upsampling method in the model can help supplement information and reduce background noise. Future research could explore alternative deformable upsampling structures that eliminate the need for traditional upsampling while still managing noise inside and outside the target.
Second, our experiments included only two recent upsampling methods, Dysample and FADE, due to their high performance as reported in prior work. For non-decoder encoder architectures, FADE’s design involves fusing pre-sampled shallow features with post-sampled features. As both Dysample and FADE require extensive hyperparameter tuning, we used their default parameters and structures for various 2D tasks. Due to the lack of 3D versions for these methods and the complexity involved in adapting their parameters, we did not replicate them in 3D for comparison. Through our experiments, we found that the combination of Dysample and DTC outperforms using transposed convolution, linear interpolation, or Dysample alone. However, since Dysample does not provide an implementation for 3D data, the effectiveness of the Dysample-DTC combination for 3D segmentation remains to be evaluated.

In addition, although DTC can improve the performance of medical image segmentation models, its effectiveness is influenced by changes in the receptive field, requiring adjustment of the hyperparameter $\lambda$. We recommend testing $\lambda$ values of 0.5, 1, 2, and 5. Notably, we observed that combining Dysample with DTC may further improve segmentation performance. Moreover, the impact of receptive field variations is more pronounced for multi-class segmentation models. In clinical applications, where it is necessary to segment each organ individually rather than merely improving overall segmentation performance, there remains considerable room for further exploration.

Finally, while DTC has demonstrated promising improvements across several classic and state-of-the-art segmentation models, its effectiveness has not yet been evaluated on a wider spectrum of architectures. Beyond segmentation, many medical imaging tasks—such as organ localization, lesion detection, and image super-resolution—also rely heavily on upsampling operations. Exploring the integration of DTC in these related tasks could provide valuable insights into its broader applicability and potential to enhance performance in diverse medical image analysis scenarios. Such investigations may uncover new opportunities for DTC to improve accuracy, robustness, and the handling of fine anatomical details in challenging clinical datasets.

\section{Conclusion}
Our extensive experiments across a wide range of 2D and 3D medical image segmentation datasets demonstrate that the deformable transposed convolution (DTC) structure can be seamlessly integrated into both classic and state-of-the-art segmentation models that rely on upsampling. By dynamically learning spatial offsets, DTC allows the network to adaptively adjust the upsampling process to better capture fine-grained structures and contextual details. This approach improves segmentation stability and accuracy compared to traditional methods, such as linear interpolation and standard transposed convolution, and shows potential benefits over recent advanced upsampling techniques. These findings suggest that DTC is a flexible and potentially useful module for enhancing the performance of diverse segmentation algorithms, particularly in challenging scenarios with complex anatomical structures or noisy backgrounds.


\bibliographystyle{cas-model2-names}

\bibliography{cas-refs}



\end{document}